\RequirePackage{silence}
\documentclass[runningheads]{llncs}
\usepackage[T1]{fontenc}
\usepackage{hyperref}
\usepackage{float}
\usepackage{graphicx}
\usepackage{caption}
\usepackage{subcaption}
\usepackage[table]{xcolor}
\usepackage{tabularx}
\usepackage{booktabs}
\usepackage{makecell}
\usepackage{array}
\usepackage{multirow}
\usepackage{makecell}
\usepackage{array}
\usepackage{graphicx}
\usepackage[table]{xcolor}

\begin{document}

\title{Exploring Learning Models for Topological Relationship Recognition from Image Data}
\titlerunning{Topological Relationship Recognition from Image Data}
\author{Saptak Das\inst{1} \and
Monidipa Das\inst{2}}
\authorrunning{S. Das and M. Das}
\institute{School of Mathematics, Indian Institute of Science Education and Research, Thiruvanantapuram, India 695551  \and
Department of Computational and Data Sciences, \\Indian Institute of Science Education and Research, Kolkata, India 741246}
\maketitle              % typeset the header of the contribution
\begin{abstract}
Figuring out how objects relate to each other—like whether they touch, overlap, stay completely separate or one sits inside another— matters a lot in fields like GIS, biomedical imaging, and robotics. Even though machine learning has come a long way, people haven’t really focused on spotting these topological relationships in images. The main roadblocks? Not enough good datasets and no clear way to measure results. So, we rolled up our sleeves and built a new dataset. It's pretty sizable: over $11,000$ labelled images showing all those essential relationships. We ran tests with some classic machine learning models—Naïve Bayes, KNN, Random Forest, SVM, and Artificial Neural Networks—and threw in some deep learning stars like VGG16 and InceptionResNetV2. For the dataset itself, we used segmentation, contour detection, and grayscale normalization to tease out solid feature vectors. The results? Deep learning methods, especially VGG16, pulled ahead, with validation accuracy hitting $89.55\%$. That’s a big jump compared to the traditional models. This shows how powerful transfer learning is for analyzing topological relationships in images, and it gives researchers a new standard to aim for in future work on spatial reasoning and topological classification.

\end{abstract}

\section{Introduction}

\begin{sloppypar}

In many fields, we see that is the relationship between objects in space is of great importance in areas like geographic information systems (GIS)~\cite{longley2015geographic}, medical imaging~\cite{litjens2017survey}, and computer vision~\cite{szegedy2015going}. We also see that the classification of topological relationships, for example, "Touch/Overlap", "Disjoint", and "Containment" which in turn improves applications from image segmentation~\cite{minaee2021image} to automated reasoning in spatial databases~\cite{shekhar2003spatial}. Despite progress in machine learning~\cite{lecun2015deep}, we still see that this is a tough problem, which we attribute to differences in object size, shape and orientation. This study’s results have large-scale implications for fields that require accurate spatial information, including robotics~\cite{siciliano2016springer}, cartography, and biomedical analysis. We present a structured data set and a study of multiple classification models, which we put forward as a contribution to the broader goal of improving automatic identification of topological relationships.

\end{sloppypar}

In spite of the fact that topological relationship classification is of great importance, there is a large gap in the present literature. Although we have seen great research in object detection~\cite{zou2019object} and image segmentation, as well as in general spatial analysis, we note that very little study has gone into the issue of the classification of topological relationships in images~\cite{papadias2001spatial}. Also, we see that there are no large-scale datasets and benchmark models, which, in turn, is a brake on progress in this field. Present solutions often use rule-based heuristics or hand-crafted feature extraction~\cite{yang2017survey}, which do not scale well to complex real-world issues. Also, we see that there is a lack of standard methods and public datasets, which, in turn, makes it hard for researchers to develop and compare new models.

The research will attempt to fill this gap by creating a dataset of around $11,000$ images, each carefully annotated to show one of three basic topological relationships: "Touch/Overlap," "Disjoint," and "Containment" This dataset will then be used as a primary resource for training and evaluating machine learning models in a field that has not received much attention so far. To take things one step further, we will apply and compare several classification methodologies: Convolutional Neural Networks (CNN)~\cite{krizhevsky2012imagenet}, Artificial Neural Networks (ANN)~\cite{haykin1994neural}, Naïve Bayes Classifier~\cite{rish2001empirical}, k-Nearest Neighbors (KNN)~\cite{altman1992introduction}, Random Forest~\cite{breiman2001random}, and Support Vector Machine (SVM)~\cite{cortes1995support}. We will use deep learning and traditional machine learning algorithms to set a benchmark for the classification of topological relationships.

This study shows a great deal of promise for many fields. In robotics, which at present is very much in need of a better understanding of object relationships, which in turn will improve grasping and manipulation tasks~\cite{bohg2014robot}. In cartography and GIS, we see that improved classification methods do, in fact, improve the interpretation of spatial data. Also in biomedical image analysis, which is a large field, we note that precise classification of which elements are overlapping or within each other can aid in the segmentation of cells, tumours, or other anatomic structures. By providing a structured data set and evaluating many machine learning models, we address a large gap in research, which we also at the same time are setting the stage for what we expect to be great advances in automated topological relationship classification. This study offers two key contributions:
\begin{itemize}
    %\item \textbf{A publicly available, large-scale dataset} dedicated to topological relationship recognition, enabling standardized training and evaluation.
    \item \textbf{A large-scale dataset:-} we have put together for the purpose of structural relationship recognition, which will be made available to the public upon acceptance of our work. 
    \item \textbf{A comparative analysis of classical and deep learning models:-} We also present a study that looks at both traditional and deep learning models and we set forth performance benchmarks, which we put forth for the research community in this relatively unexplored but very important field.
\end{itemize}

\section{Related Works}

Early, a lot of methods used rule-based heuristics and manually created features~\cite{egenhofer1991topological}, which produced some insight but also had issues with scale and reliability in complex real-world settings.

Also, to the base of what we see today in spatial relationships, we have the 9 intersection model put forth by Egenhofer and Herring~\cite{egenhofer1991topological}, which puts forth basic relationships like Disjoint, Meet, Overlap, and Contain. This framework has seen wide use in spatial reasoning systems and in the optimization of spatial queries which includes in the index approaches for spatiotemporal data~\cite{papadias2001indexing}.

The growth of machine learning, in particular deep learning, has transformed areas like object detection and scene understanding. Convolutional Neural Networks (CNNs) have reported great results in many vision tasks~\cite{krizhevsky2012imagenet,he2016deep}. But their use in topological relationship classification is still at an early stage. Some research has reported on the use of CNNs for spatial relationship reasoning, which in large part has been to look at geometric proximity and function relationship rather than pure topological semantics~\cite{zhang2019spatial}.

Also, a large issue in this field is the lack of available data sets that properly tag topological relationships. While we do have datasets like Visual Genome~\cite{krishna2017visual}, which provide in-depth annotation for spatial interactions. We see that they tend to put more into the context and function relationships, which, in turn, do not fully cover formal topological structures.

We present our work, which fills in these gaps by introducing a new large-scale dataset that we have annotated for the most important topological relationships: Touch/Overlap, Disjoint, and Containment. Also, we evaluate and benchmark multiple machine learning and deep learning models, which we do to encourage more research and development in automatic topological relationship classification.

\section{Methodology}
\subsection{Image Processing and Image Labeling:}

\textbf{Image Processing:} Image analysis was performed to prepare the dataset for segmentation and analysis. We resized all input images to the same resolution, which in turn made the dataset consistent. We used K-means clustering with \( K = 3 \) to divide each image into distinct areas by pixel intensity. Also, we converted the resulting segments to grayscale and then applied Otsu’s thresholding for binarization. We did morphological operations, which included a closing function with a rectangular structuring element to improve the structure of the segments and also to remove noise. These pre-processing steps, in turn, improved the accuracy of the contour detection, which, in turn, increased the reliability of the results from our spatial relationships analysis.
\begin{figure}[htbp]
    \centering
    \includegraphics[width=\textwidth]{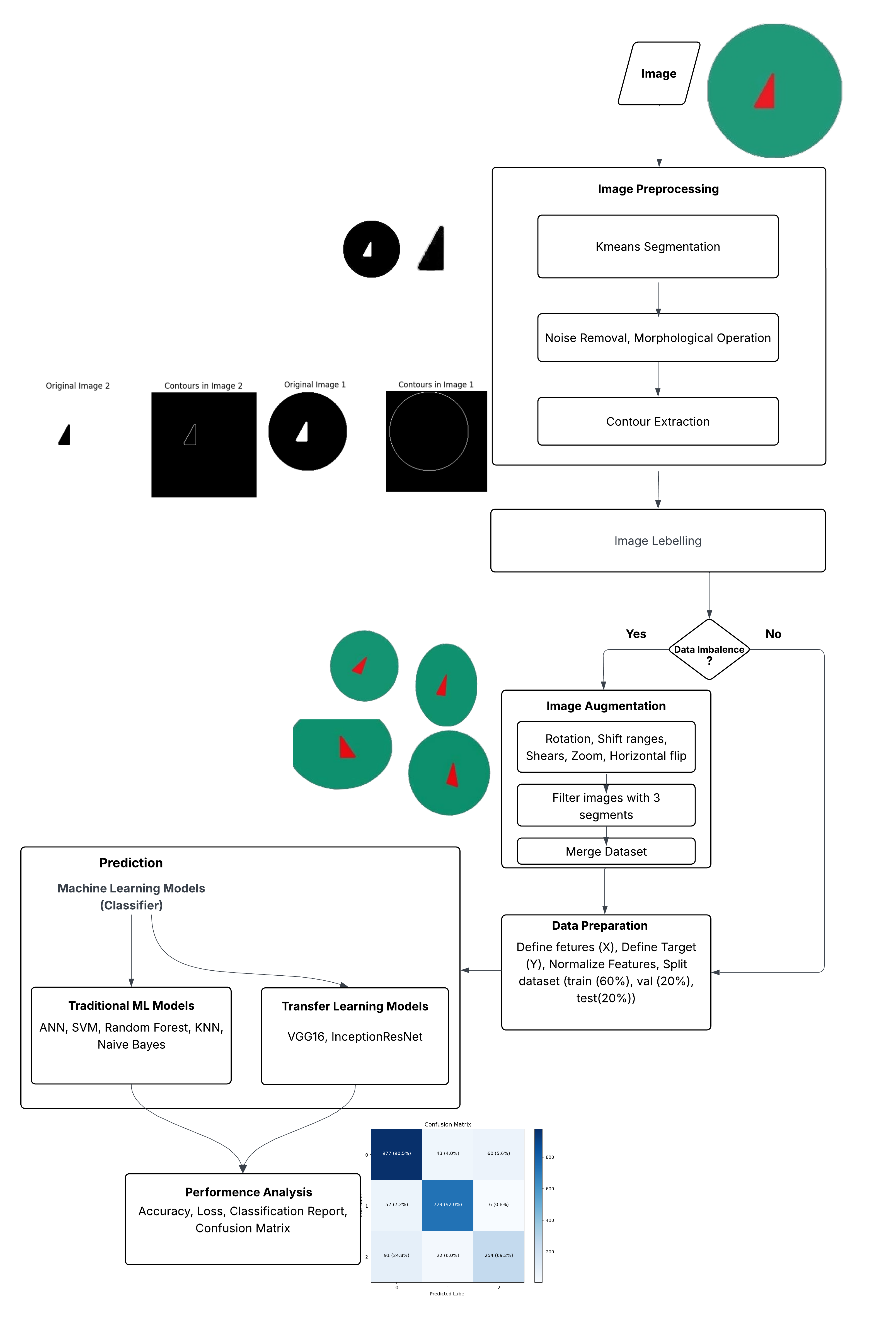} 
    \caption{Methodology Flowchart Diagram}
    \label{fig:flowchart}
\end{figure}

% \textbf{Image Labeling:} Image labeling was achieved through spatial analysis of the contours extracted from segmented image regions. For each image, two primary segments were selected, and their contours were extracted using the Canny edge detection algorithm followed by contour tracing via the \texttt{cv2.findContours()} function. The spatial relationship between the two segments was determined based on the degree of contour point intersection. Images were then categorized into one of three classes: \textit{Containment}, \textit{Disjoint}, or \textit{Touch/Overlap}. 
\noindent\textbf{Image Labelling:} Image labelling was achieved through spatial analysis of the contours extracted from segmented image regions. For each image, two primary segments were selected to be identified as spatial objects, and their contours were extracted using the Canny edge detection algorithm followed by contour tracing via the \texttt{cv2.findContours()} function. The spatial relationship between the two segments was determined based on the degree of proximity between two sets of contour points. Images were then categorized into one of three classes: \textit{Containment}, \textit{Disjoint}, or \textit{Touch/Overlap}. In particular, an image was annotated with the label `Containment' if all the contour points of one of the objects were found to be neighbours of the contour points of the other object. In case the neighbourhood relationship was satisfied partially, the image was annotated with the label `Touch/Overlap'. Otherwise, (when none of the points on object contours satisfied the neighbourhood relationship), the image was annotated as `Disjoint'. Regarding neighbourhood relationship, we considered rook case, where the neighbourhood of any point $p(x,y)$ is defined as: $$\mathcal{N}_{p}=\left\lbrace(x,y+1),(x,y-1),(x-1,y),(x+1,y)\right\rbrace$$

These categorical labels/annotations were subsequently mapped to numerical values (2 for Containment, 1 for Disjoint, and 0 for Touch/Overlap) to support machine learning applications. Additionally, pixel-level grayscale values were extracted from resized (24×24) versions of each image to form a structured dataset, which was exported to an Excel file for further processing. 

In this study we present in figure \ref{fig:flowchart} the overall workflow. It details the key steps we took in the analysis of topological relationships from image data, which began with image pre-processing and labelling, then we moved on to feature vector generation, dataset partitioning, model training (classical and deep learning), hyperparameter tuning, and performance evaluation. 

\subsection{Machine Learning and Deep Learning Models:}

\subsubsection{Dataset Description:}
The dataset, in Excel format, contains feature vectors obtained from image data. Each instance is classified into one of three topological categories: Overlap, Disjoint, or Containment. The non-informative columns Image Name and Case are removed from the dataset. We normalized feature values to the $[0, 1]$ range by dividing by $255.0$, which in turn improves the training accuracy and convergence speed of neural networks.

This dataset was split into three subsets: the training set was 60\% of the data, while both the validation set and the testing set were 20\% of the data.
\vspace{0.0cm}

\subsubsection{Model Summary and Configuration:}

Model-specific differences in training configurations:

\begin{table}[!t]
\caption{Summary of Machine Learning and Deep Learning Models}
\centering
\small 
\renewcommand{\arraystretch}{1.2}
\begin{tabularx}{\textwidth}{|>{\raggedright\arraybackslash}p{3cm}|>{\raggedright\arraybackslash}X|>{\raggedright\arraybackslash}p{5cm}|}
\hline
\centering \textbf{Model} & \textbf{Key Characteristics} & \textbf{Best Hyperparameters} \\
\hline
\makecell{ANN} & Fully connected feedforward neural network (MLP); suitable for multi-class classification & 
\begin{itemize}
    \item Learning rate: $0.03$
    \item Activation: ELU
    \item Hidden layers: $2$
    \item Neurons per layer: $105$
\end{itemize} \\
\hline
\makecell{SVM} & High-dimensional classifier with kernel tricks & 
\begin{itemize}
    \item C: $10$
    \item Kernel: rbf
    \item Gamma: scale
    \item Class weight: balanced
    \item Best Cross Validation $0.70$
\end{itemize} \\
\hline
\makecell {Random\\Forest} & Ensemble decision trees; robust and interpretable & 
\begin{itemize}
    \item n estimators: $200$
    \item mMdepth: None
    \item Min samples\_split: $2$
    \item Max features: log2
    \item Bootstrap: False
    \item Class weight: none
    \item Cross validation: $0.72$
\end{itemize} \\
\hline
\makecell{KNN} & Distance-based instance learner; effective on non-linear data & 
\begin{itemize}
    \item n\_neighbors: $7$
    \item Weights: distance
    \item Metric: manhattan
    \item cross validation: $0.60$
\end{itemize} \\
\hline 
\makecell {Naïve\\ Bayes} & Gaussian Naïve Bayes; efficient for high-dimensional, small datasets & 
\begin{itemize}
    \item No tunable parameters
\end{itemize} \\
\hline
\makecell {CNN\\(VGG16)} & Transfer learning with VGG16; Dense softmax head &
\begin{itemize}
    %\item VGG16 layers frozen
    \item No tuning on base layers
\end{itemize} \\
\hline  
\makecell {CNN\\(InceptionResNetV2)} & Deep CNN with Flatten and softmax output layer & 
\begin{itemize}
    \item No tuning on base layers
\end{itemize} \\
\hline
\end{tabularx}
\label{tab:ml_dl_summary}
\end{table}

All models were trained with Adam optimizer along with categorical cross-entropy loss for $100$ epochs. optimization was performed using Keras Tuner with Random Search for neural network-based models and RandomizedSearchCV with stratified 5-fold cross-validation for classical machine learning models.

\begin{itemize}
  \item \textbf{CNN (VGG16):} Utilized transfer learning with frozen base layers. Data augmentation was applied.
  \item \textbf{CNN (InceptionResNetV2):} Employed early stopping based on validation loss. The base layers were not tuned.
  \item \textbf{Naïve Bayes:} Trained directly without any parameter tuning or validation steps.
\end{itemize}
\subsubsection{Model Evaluation:}

All models were evaluated separately on the \textbf{training}, \textbf{validation}, and \textbf{test} datasets using \textbf{classification accuracy} as the primary performance metric. For the Artificial Neural Network (ANN), Support Vector Machine (SVM), Random Forest, K-Nearest Neighbors (KNN), Naïve Bayes models and the CNN-based models using \textbf{VGG16} and \textbf{InceptionResNetV2} with transfer learning, \textbf{confusion matrices} were computed to analyze class-wise prediction accuracy and to visualize misclassifications. Additionally, \textbf{classification reports} were generated for these models to provide precision, recall, and F1-scores. In the case of InceptionResNetV2, \textbf{early stopping} was employed to prevent overfitting.

\vspace{0.0cm}

%\textbf{Confusion Matrix:}
\begin{figure}[H]
    \centering
    \begin{subfigure}[b]{0.32\textwidth}
        \includegraphics[width=\textwidth]{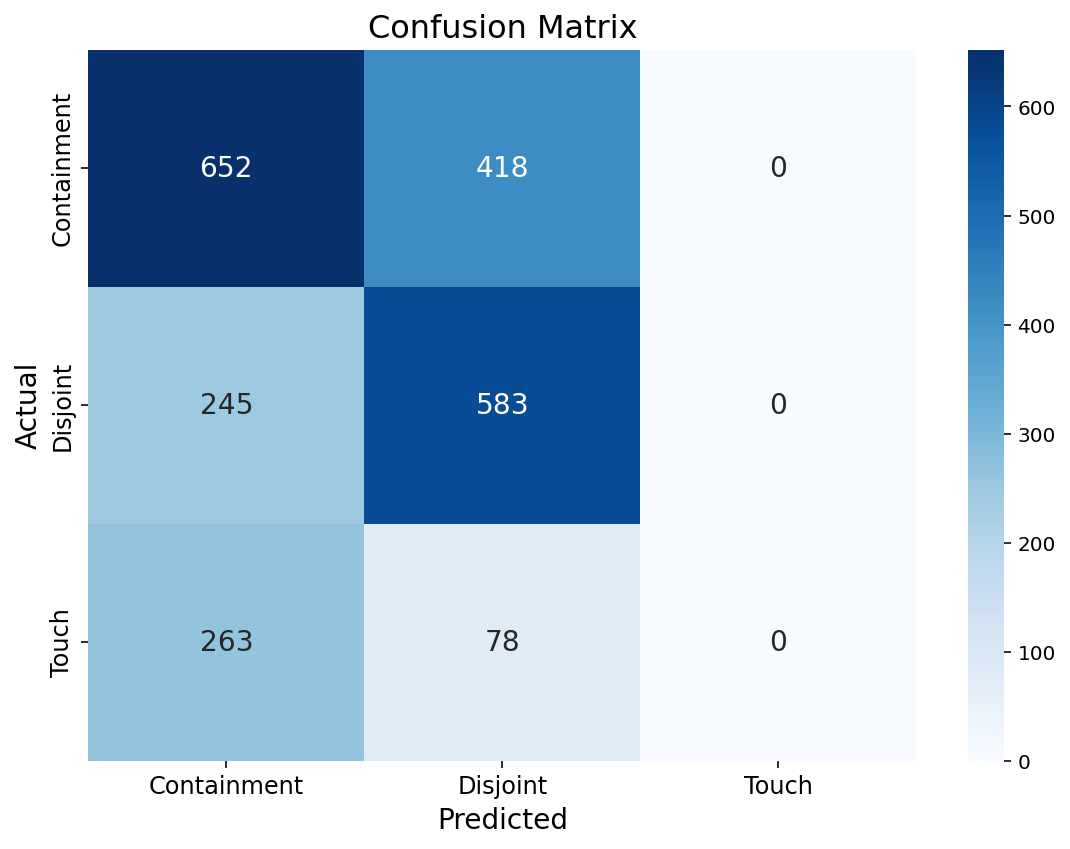}
        \caption{ANN}
    \end{subfigure}
    \hfill
    \begin{subfigure}[b]{0.32\textwidth}
        \includegraphics[width=\textwidth]{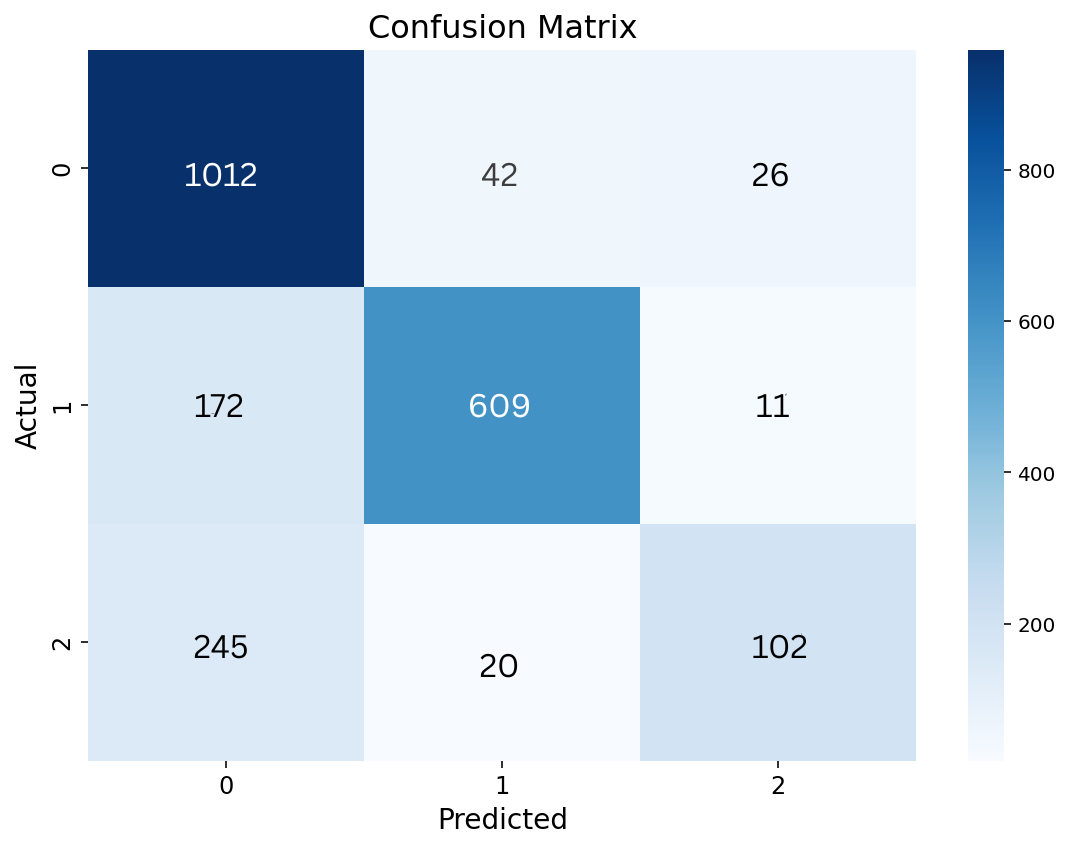}
        \caption{InceptionResNetV2}
    \end{subfigure}
   
    \begin{subfigure}[b]{0.32\textwidth}
        \includegraphics[width=\textwidth]{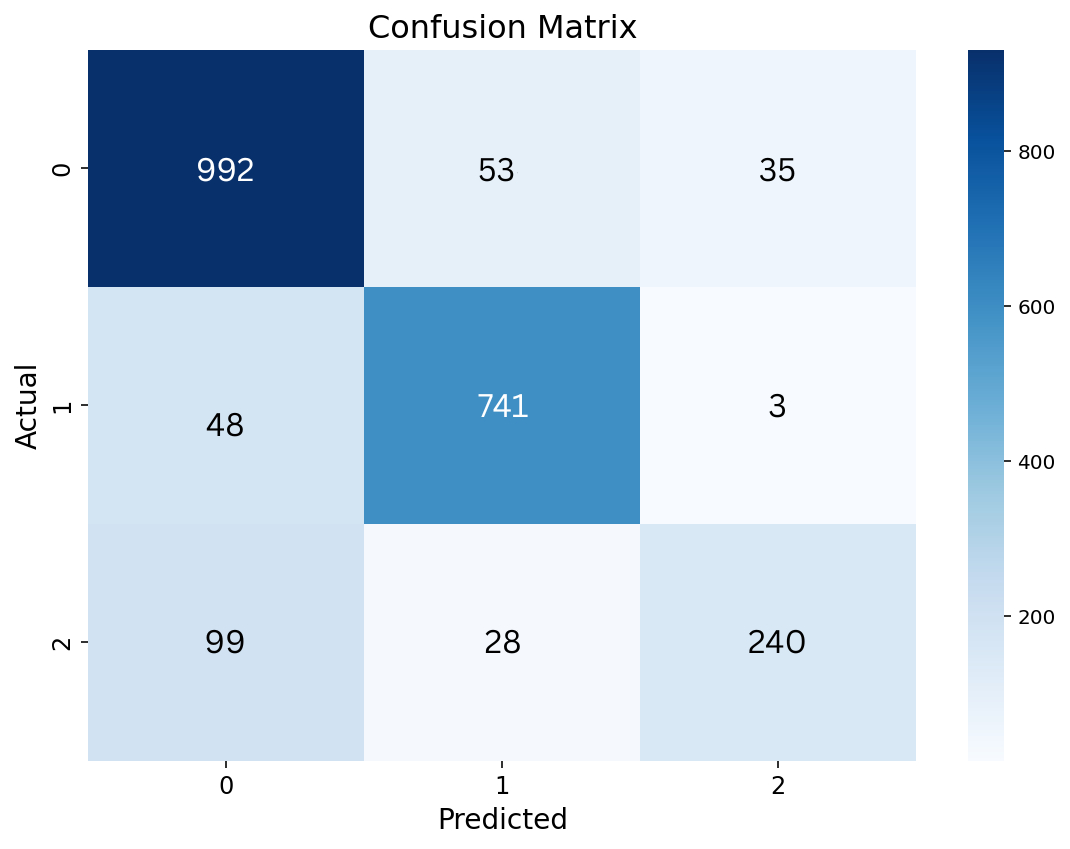}
        \caption{VGG16}
    \end{subfigure}
    \hfill
    \begin{subfigure}[b]{0.32\textwidth}
        \includegraphics[width=\textwidth]{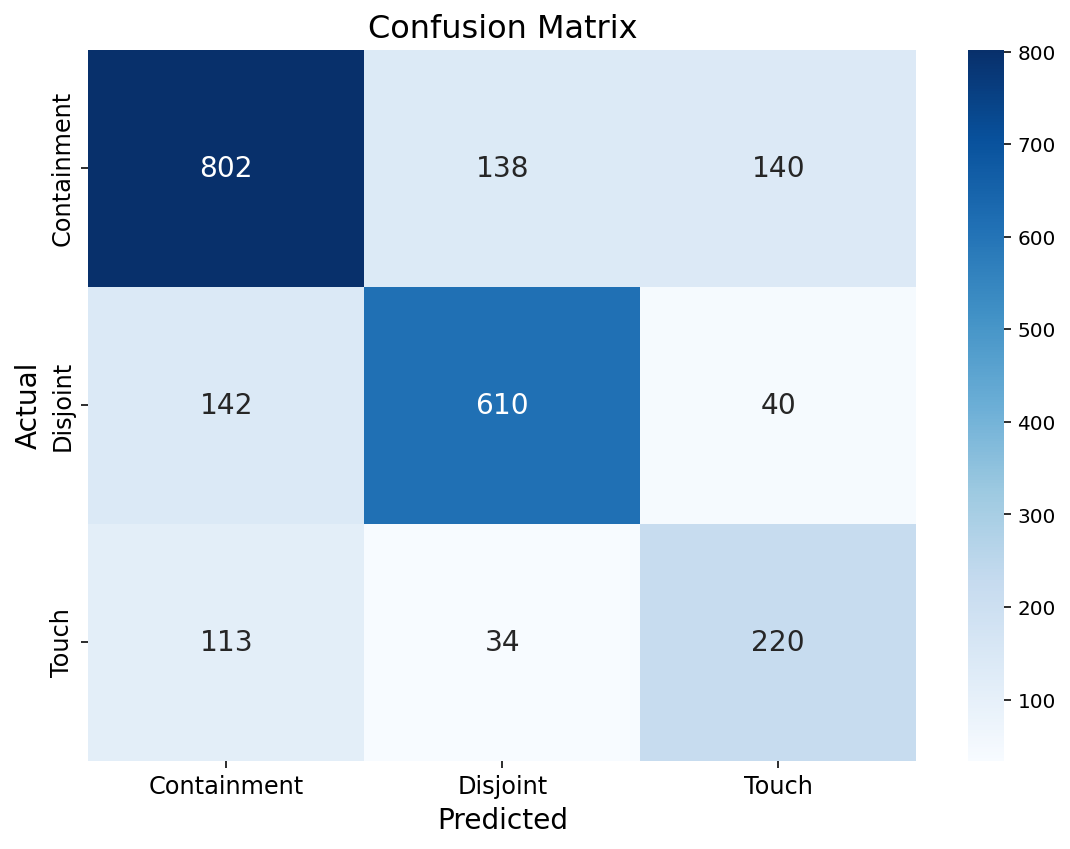}
        \caption{SVM}
    \end{subfigure}

    \begin{subfigure}[b]{0.32\textwidth}
        \includegraphics[width=\textwidth]{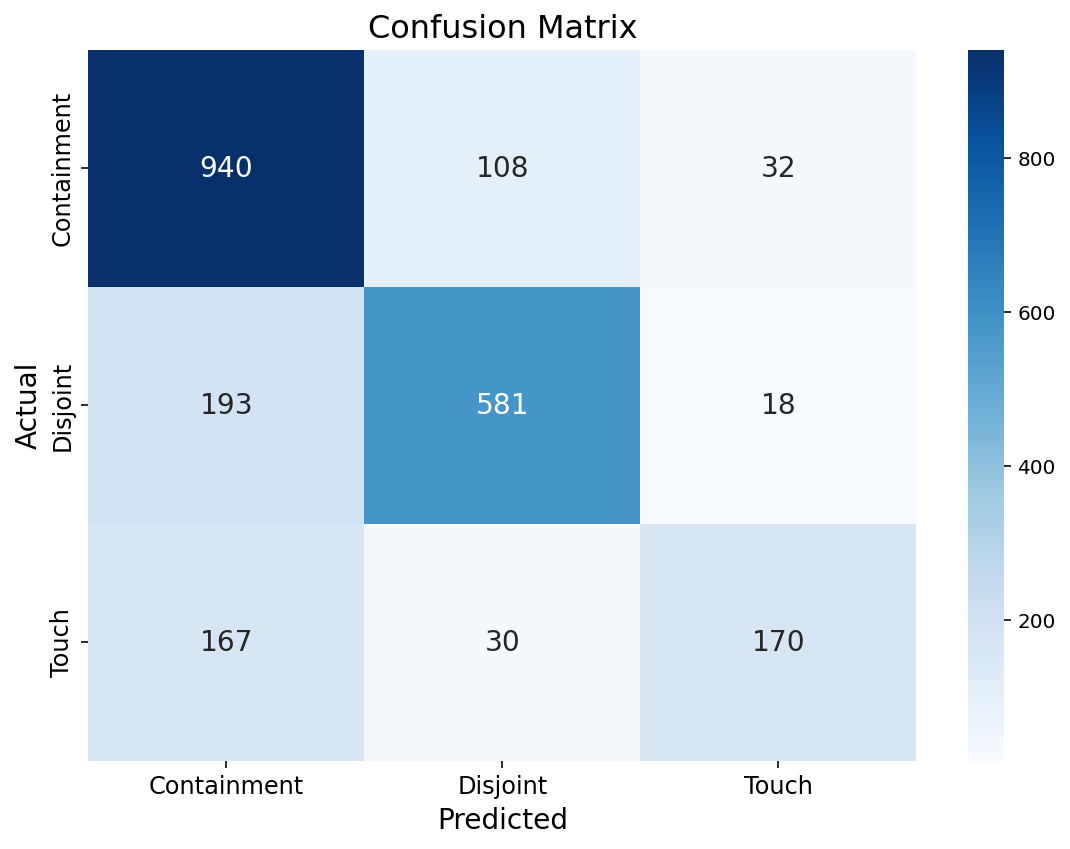}
        \caption{Random Forest}
    \end{subfigure}
    \hfill
    \begin{subfigure}[b]{0.32\textwidth}
        \includegraphics[width=\textwidth]{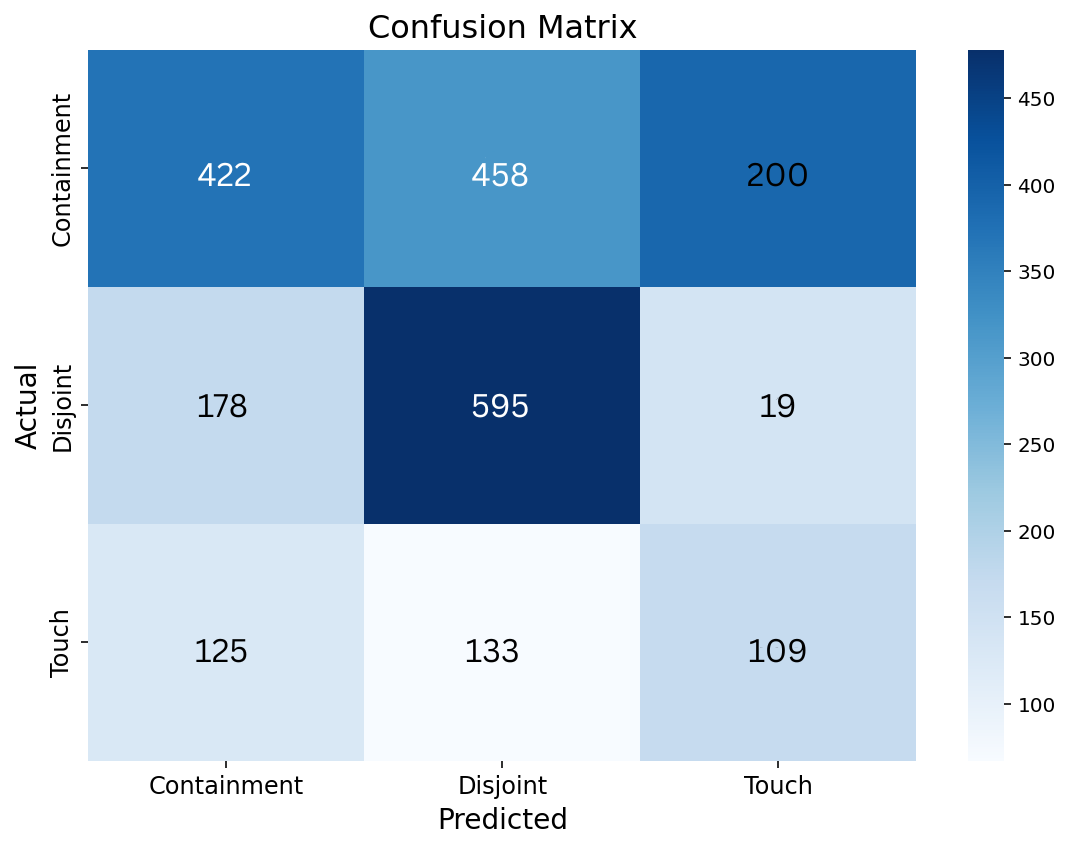}
        \caption{Naïve Bayes}
    \end{subfigure}

\end{figure}

\begin{figure}[H]\ContinuedFloat
    \centering
    \begin{subfigure}[b]{0.32\textwidth}
        \centering
        \includegraphics[width=\textwidth]{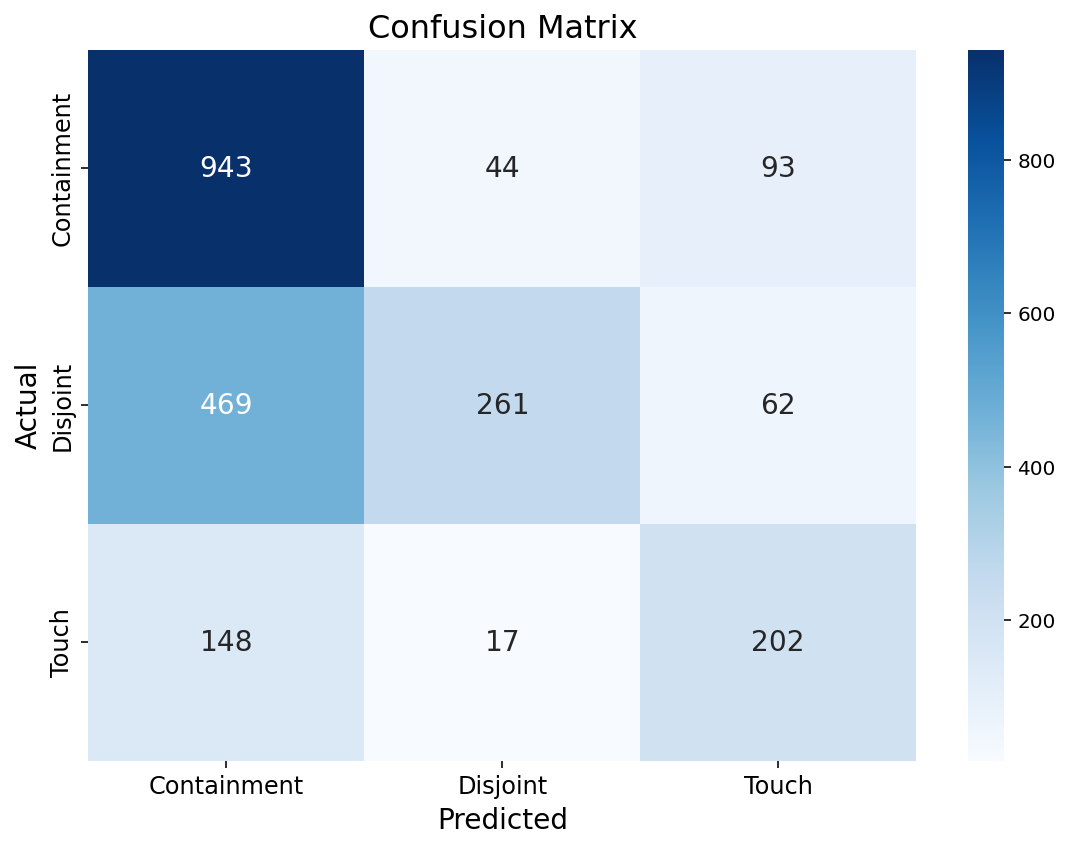}
        \caption{KNN}
    \end{subfigure}

    \caption{Confusion Matrix for different classification models}
    \label{fig:Confusion Matrix}
\end{figure}
\section{Experimental Study}
In this report, we present the results of a study that we designed to compare various machine learning and deep learning models on the given dataset. We looked at how each model performed during training, testing and validation. We looked at traditional and deep learning approaches to determine which performs better on the classification task and what issues each has. 
\vspace{0.5cm}
\newline 
\textbf{Analysis:}
The tables that follow report the performance of each model in terms of training, testing and validation accuracy and also the model classification report.

\begin{table}[!h]
\caption{Performance analysis of various classification models, which is based on precision, recall, and F1-score.}
\centering
\renewcommand\theadfont{\bfseries}
\rowcolors{2}{white}{white!10}
\begin{tabular}{|>{\raggedright\arraybackslash}p{1.4cm}|>{\raggedright\arraybackslash}p{5.4cm}|>{\centering\arraybackslash}p{1.5cm}|>{\centering\arraybackslash}p{1.5cm}|>{\centering\arraybackslash}p{1.5cm}|}
\hline
\rowcolor{white}
\makecell{\textbf{\textcolor{black}{Model}}\\\textbf{\textcolor{black}{Type}}} & \centering \textbf{Model Name} & \textbf{Pression} & \textbf{Recall} & \textbf{F1-score} \\
\hline
\makecell {ANN} & Without Dropout Ratio (With Hyperparameter Tuning) & 0.37 & 0.45 & 0.41 \\
\hline
\makecell {SVM} & SVM with Hyperparameter Tuning (RandomSearch CV) & 0.70 & 0.70 & 0.70 \\
\hline
\makecell {Random\\Forest} & RF with Hyperparameter Tuning (RandomSearch CV) & 0.77 & 0.69 & 0.71 \\
\hline
\makecell {KNN} & KNN with Hyperparameter Tuning (RandomSearch CV) & 0.66 & 0.58 & 0.58 \\
\hline
\makecell {Naïve\\ Bayes} & NB without HyperParameter Tuning (RandomSearch CV) & 0.47 & 0.48 & 0.48 \\
\hline
\makecell {\textbf{CNN}} & \textbf{VGG16 Model} & \textbf{0.88} & \textbf{0.84} & \textbf{0.85} \\
\hline
\makecell {\textbf{CNN}} & \textbf{InceptionResNetV2} & \textbf{0.78} & \textbf{0.66} & \textbf{0.68} \\
\hline
\end{tabular}
\end{table}
\begin{table}[!t]
\caption{Performance Analysis}
\centering
\renewcommand{\arraystretch}{1.5}
\resizebox{\textwidth}{!}{%
\begin{tabular}{|c|p{5.2cm}|c|c|c|}
\hline
\rowcolor{white}
\makecell{\textbf{\textcolor{black}{Model}}\\\textbf{\textcolor{black}{Type}}} & \centering\arraybackslash\textbf{\textcolor{black}{Model Name}} & 
\makecell{\textbf{\textcolor{black}{Training}}\\\textbf{\textcolor{black}{Accuracy (\%)}}} & 
\makecell{\textbf{\textcolor{black}{Validation}}\\\textbf{\textcolor{black}{Accuracy (\%)}}} & 
\makecell{\textbf{\textcolor{black}{Test}}\\\textbf{\textcolor{black}{Accuracy (\%)}}} \\
\hline

ANN & Without Dropout Ratio \newline (With Hyperparameter Tuning) & 55.11 & 54.19 & 56.19 \\
\hline
SVM & SVM with Hyperparameter Tuning \newline (RandomSearch CV) & 96.05 & 73.11 & 72.89 \\
\hline
\makecell {Random\\Forest} & RF with Hyperparameter Tuning \newline (RandomSearch CV) & 99.67 & 75.97 & 75.52 \\
\hline
KNN & KNN with Hyperparameter Tuning \newline (RandomSearch CV) & 99.67 & 63.91 & 62.8 \\
\hline
\makecell {Naïve\\ Bayes} & NB without HyperParameter Tuning \newline (RandomSearch CV) & 51.2 & 46.45 & 50.29 \\
\hline
\textbf{CNN} & \textbf{VGG16 Model} & \textbf{90.89} & \textbf{88.74} & \textbf{87.54} \\
\hline
\textbf{CNN} & \textbf{InceptionResNetV2} & \textbf{84.59} & \textbf{84.06} & \textbf{83.16} \\
\hline
\end{tabular}%
}
\label{tab:tab3}
\end{table}

\vspace{1cm}

From the Table~\ref{tab:tab3}, it is noted that:
\begin{itemize}
    \item \textbf{CNN-based models (VGG16 and InceptionResNetV2)} report very high accuracy across all splits. In particular, VGG16 does which of $89.55\%$ in validation accuracy, which is a sign of great generalization and feature extraction ability.
    \item \textbf{Random Forest and SVM} also perform very well, with validation accuracies in the $75\%$ range. Also, for these models, we see that hyperparameter tuning via RandomSearch CV improves performance.
    \item \textbf{KNN}, although it achieves perfect results in training, it does still suffer from overfitting, which we see in a large drop of performance for testing and validation accuracy. 
    \item \textbf{Naïve Bayes and ANN (without dropout)} perform very weakly; we think this is because they are not able to model the dataset’s complexity well enough.
    \item We did tune all classical models (except Naïve Bayes) with RandomSearch CV, ensuring a fair comparison.
\end{itemize}

In our experiments, we note that the success of deep learning approaches is very much demonstrated; in particular, we see that pre-trained CNN architectures like VGG16 and Inception-ResNetV2 do very well on large-scale classification tasks. Also, we find that in terms of validation accuracy and generalization these models outperform traditional machine learning models. 

While models like Random Forest and SVM do well when tuned, they still
can’t match the performance of CNNs in image-based data. Also, we see that simpler models like Naïve Bayes and poorly tuned ANN models do not perform well because they cannot capture enough features.

\section{Conclusion and future scope:}
In this study, we present a foundation for future research, including the development and fair comparison of new algorithms for topological relationship recognition. We report on large-scale applications in GIS, biomedical imaging, and robotics that depend heavily on accurate spatial relationship analysis. In the coming research, we will look at more complex architectures, multimodal data and domain-specific improvements, which in turn will improve performance and practical application. Models that include Naïve Bayes and underperforming ANNs showed that model selection and tuning of hyperparameters are of great importance in this field.

We present in this study a base that we put forth for future research, which also includes the development and fair comparison of new algorithms for topological relationship recognition. Our work has large-scale applications in GIS, biomedical imaging and robotics, which are very much dependent on accurate spatial relationship analysis. In the coming research, we will look into more complex architectures, multimodal data and domain-specific improvements, which in turn will improve performance and practical application.

In the future, we will also expand our research into medical imaging, including MRI and CT data. It is in these that we see structural and textural features, which may support the distinction between normal and cancerous tissues. In the future, this work may contribute to the development of an interpretive decision support tool for use in histopathological analysis and clinical diagnosis.

\end{document}